# A novel method for iris recognition using BP neural network and parallel computing by the aid of GPUs (Graphics Processing Units)

F.Hosseini, H.Ebrahimpour, S.Askari

Department of Computer Engineering, University of Kashan, P.O . BOX 51167-87317,Ravand Blvd,Kashan,Iran , E-mail: Farahnazhosseini@grad.kashanu.ac.ir

Department of Computer Engineering, University of Kashan, P.O . BOX 51167-87317,Ravand Blvd,Kashan,Iran , E-mail: Ebrahimpour@kashanu.ac.ir

Department of Computer Engineering, Islamic Azad University of Arak Science & Research Branch, Arak ,Iran, E-mail: Samaneh.askari@yahoo.com

**Abstract.** In this paper, we seek a new method in designing an iris recognition system. In this method, first the Haar wavelet features are extracted from iris images. The advantage of using these features is the high-speed extraction, as well as being unique to each iris. Then the back propagation neural network (BPNN) is used as a classifier. In this system, the BPNN parallel algorithms and their implementation on GPUs have been used by the aid of CUDA in order to speed up the learning process. Finally, the system performance and the speeding outcomes in a way that this algorithm is done in series are presented.

**Keywords:** Iris recognition system, Haar wavelet, Graphics Processing Units (GPUs), neural network BP, CUDA.

## 1 Introduction

To find out the human identity has been a long-term goal of humanity itself. Since the technologies and services are rapidly developing, an urgent need is required to identify a person. Examples include passport, computer login and in general, the security system's control. The requirements for this identification are the speed and increased reliability. Biometric as a model for human identification is one of the interesting domains to the researchers intended to increase the speed and security. Different biometric features offer various degrees of reliability and efficiency. Iris of a person is an internal organ in the eye, yet easily visible from one meter distance thus is a perfect biometric identification system. Iris recognition is widely accepted as one of the best biometric identification methods throughout the world.

It has several advantages: First, its pattern variability is high among different individuals; therefore it meets the needs for unique identification. Second, the iris remains stable throughout life. Third, the iris image is almost insensitive to angled radiation and light, and every change in the angle leads only to some relative transferences. Fourth, according to Sandipan p (2009) the ease of eyes' localization in the face and the distinctive shape of iris ring add to its reliability and make its isolation more accurate. Several algorithms have been developed for the iris recognition system. All algorithms include a variety of steps such as: to obtain an image to localize the iris, to extract the features, to match and classify them. The first automatic iris recognition system was proposed by Daugman (1993). He used the Gabor filters to extract the features. While Daugman was improving his algorithm [3] several researchers have also been working on iris recognition. Wildes (1997) used a laplacian pyramid to represent the iris tissue. Boles and Boashash (1998) employed one-dimensional wavelet transforms at different resolution of concentric circles on an iris image. By using two-dimensional Haar wavelet, Lim (2001) decomposed the iris image into 4 levels, and he used a competitive learning neural network (LVQ) as a classifier. In all the above-mentioned methods, the serial algorithm has been used for iris recognition. In this paper, by using parallel algorithms and GPUs, we present a method that has increased the response speed substantially. The database used in this article is CASIA V.3.The rest of the paper is organized as the following: In section 2, the iris recognition system will be explained. In section 3, the results obtained from this implementation will be presented and finally, section 5 includes the conclusion along with the instructions for the future works.

## 2. Iris Recognition System

An iris recognition system generally consists of the following three parts: pre-processing and normalizing the iris, extracting the features, and classifying the extracted features.

### 2-1 Iris Pre-processing

Iris Pre-processing includes localization and normalizing the iris which basically consist of two main operations: One is to recognize the eyelashes and the other is to recognize the boundaries. The first step is to extract the circular edge of iris by removing Noisy areas. Eyelids and eyelashes cover the upper and lower portions of iris. Therefore, these areas should be segmented. The second step involves recognizing the inner and outer boundaries of iris. One is in the passage area of the iris and sclera and the other is in the iris and pupil. For this reason, canny edge detection for both horizontal and vertical direction is suggested by Wildes (1999). We have used the circular Hough transform method to localize the iris and the linear Hough transforms to separate the eyelids. Hough transform is a standard algorithm which can be used to determine the simple geometric shapes such as circles and lines. If we use the Hough transform first for the edge of pupil-sclera and then for the edge of iris-pupil, more accurate results will be obtained. Thus, the canny edge detection is first used to create edges and then the circles surrounding the iris space are obtained by using the Hough transform method. If the maximum Hough space is less than the threshold, then the block will not be displayed by the eyelids. It's easier to separate the eyelashes by thresholding because they are darker compared with other organs in the eye.

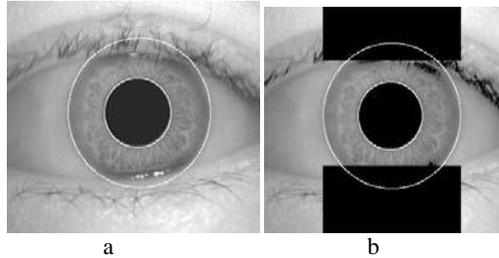

a) Iris image after detecting the edges by Hough algorithm.
b) Iris Image after removing the Noises.
**Fig.1.** Iris Localization.

Daugman suggested the normal Cartesian coordinates for polar conversion that maps every pixel in the space to a pair of polar coordinates $(r,\theta)$ which r and $\theta$ are in the range of $[0,1]$ and $[0,2\pi]$ respectively. This mapping can be formulated as following:

$$I(x(r,\theta)) \to I(r,\theta) \qquad (1)$$
$$x(r,\theta) = (1-r)x_p(\theta) + rx_1(\theta) \qquad (2)$$
$$y(r,\theta) = (1-r)y_p(\theta) + ry_1(\theta) \qquad (3)$$

Where $(x_i, y_i)$, $(x_p, y_p)$, $(r,\theta)$, (x,y) and I(x,y) show the iris area, Cartesian coordinates, polar coordinates and the coordinates of the pupil and iris boundaries respectively in    direction. According to Pandara and Chandra (2009) The rotational incompatibilities are not considered in this display. An example of this polar transform can be seen in Fig2.

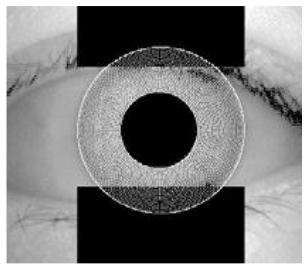

a) Normalized iris form

b) The Polar form of iris

**Fig.2.** Example of polar transform for an iris.

## 2-2 Feature extraction

One of the key issues in identification system speed is the size of feature vector extracted from every iris. To reduce the size of iris feature's vector, an algorithm must be applied that does not miss the general and local information obtained from iris. We have used Haar wavelet to extract the iris features vector. This wavelet is one of the fastest wavelet's transformations. In this paper we have used the two-level Haar wavelet, i.e. the image of iris is decomposed and the coefficients that represent the iris pattern are obtained. The features that reveal the additional information are also deleted. As can be seen in Figure 3, in 2-level transform, one image of iris is decomposed into 7 sections, one of which is only (the above picture, left corner) is extracted as the main feature and the other sections are deleted as the additional information.

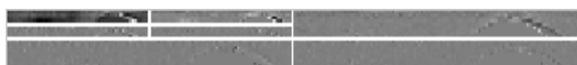

**Fig.3.** Haar transform extracted from the iris area in 2-level.

## 2.3. Classification

We have used the Back Propagation Neural Network (BPNN) to classify the feature vectors. According to Shylaja (2011) BPNN is a systematic method for training multi-layer artificial neural network. This network provides a computationally effective way to change the weights in a feed forward network. The basic structure of BPNN consists of an input layer, at least one hidden layer and output layer. The neural network works with regulating the weight values to reduce the errors between the actual output and the resulting output.

BPNN pseudo-code algorithm used in this section is presented in Figure 4. As it can be seen, for a BP network with a middle layer, updating the weight for a given input has four stages, assuming that the network has n input neurons, h neurons in the hidden layers and o neurons in the outer layers. These four stages are:

$$
\begin{aligned}
&\textit{While MSE is not satisfied} \\
&\quad \textit{for all input vectors } x \\
&\qquad \textit{Flayer}(x); \\
&\qquad \textit{Slayer}(x); \\
&\qquad \textit{Backpro\_Flayer}(x); \\
&\qquad \textit{Backpro\_layer}(x); \\
&\quad \textit{recompute MSE}
\end{aligned}
$$

**Fig.4.** BPNN algorithm

In the first stage, the values of middle layer should be specified for every input. In the second stage, the outer layer's values are obtained from the middle layer. In the third stage, considering the errors resulted from the output layer, the weight values between the hidden layers and output layers must be updated and finally, in the fourth stage, the weight values between the first layer and the middle layer are updated. These operations are repeated successively in order to keep the general error rate at an acceptable number.

## 3. Parallel processing using CUDA

### 3.1. An Introduction to GPUs

Today, the GPUs installed on graphic cards, have an exceptional processing power comparing to the central processors. This issue has enhanced these processors' application in areas beyond the computer games. Modern graphic processors with their parallel architecture are considered very fast processors. Graphics processing units or GPUs are specific tools that are used for graphic rendering (to create natural-looking images) in personal computers, workstations, or on gaming consoles.
CUDA (Compute Unified Device Architecture) is a parallel computing architecture that was presented by the NIVIDIA Company in 2006 in order to carry out massive parallel computations with high efficiency on the GPUs developed on this company. CUDA is the computational engine of GPUs made in NVIDIA which is available in the form of different functions in programming languages to the software developers.
CUDA comes with a software environment that allows the software developers to perform their programming in C language and to run on the GPUs.
Each CUDA program consists of two parts: Host and Device. The Host is a program that is run sequentially on CPU, and the device is a program that is run on the GPU cores in a parallel way.
From software perspective, each parallel program could be considered to consist of a number of strings. These strings are light-weight processors each of which performs an independent operation. A number of dependent strings form a block and a number of blocks form a grid.
There are various types of memories in CPUs. Each string has a specific local memory of its own, each block has a shared memory which the inner strings has access to this memory. There is a global memory that all the strings have access to it. Besides, there is another type of memory called 'texture memory' which like the global memory, all the strings have access to it but the addressing mode is different and is used for specific data.
In the Host, the number of strings or in other words the number of light-weighted processors which are to be run on GPU cores should be specified. The Device code is run based on the number of defined strings in Host. Each string can find its position by the specified functions in CUDA and can do its task considering its position. Finally, the computed results should be returned to the main memory. GPUs are on excellent tool for implementing the image processing algorithms since many operators that act on the image are local and should be applied on all pixels. Thus, by considering one string for each pixel (in case that the number of required strings could be defined), the computation time could be significantly reduced. Yang (2008) has implemented a number most popular images by CUDA. In addition, in one of our previous studies, CUDA was used for spatial image processing and or Gray (2008) has used CUDA in order to determine the course of action.

### 3.2. BPNN paralleling by CUDA
Since the iterations in learning algorithm of BP network are interdependent, these iterations could not be done in a parallel way, thus only the operations of iteration could be parallel.

In this paper, all the four steps of BPNN in iteration are made parallel by CUDA. In the first stage, the values of middle layer should be specified for inputs. For this purpose, we use a block with h string. Initially, each string finds out which neuron a hidden layer is allocated to. Then, it updates the values of neuron according to values of input n neuron. Pseudo-code of these strings is given in Figure 5.

$$\begin{aligned}
&Begin\\
&\quad i : the\ thread\ Id\\
&\quad sum \leftarrow 0\\
&\quad For\ all\ nerons\ k\ at\ input\ layers\\
&\quad\quad sum \leftarrow sum + W_1(k,i) * value\ of\ neron\ k\\
&\quad hidden[i] \leftarrow sigmoid[sum]\\
&End
\end{aligned}$$

**Fig.5.** Pseudo-code of parallel BPNN algorithm in the first layer

Using this method, the entire values of the middle layers of neurons are calculated simultaneously.
Similarly, the rest of the process is implemented by CUDA.

## 4. Results
To test the system, the iris images of 100 people were used from CASIA3 database. 5 iris images were selected for every person's iris. Thus, the whole selected database includes 500 iris images.

The dimensions of extracted iris area after pre-processing and normalizing iris images (in the way it was mentioned in the previous section) are $20 \times 480$. Thus, the extracted features of Haar Wavelet from the images are obtained in $5 \times 60$ dimension by two levels. After ordering the achieved values in one dimension, finally a vector of 30 is obtained as a feature from every iris image.

One BPNN network of 300 neurons in input layer and 7 neurons in output layer is used in the classification of the obtained feature. The output of this network is a 7 bit binary data that represents a person to whom the input feature vector belongs. Of the total data, 80 and 20 per cent were considered as the training data and test data respectively.

To test the system, different neuron numbers were used in the BPNN middle layer. The average precision obtained for every 5 times system training is mentioned for each of them in the table1:

**Table 1.**

| The gained precision by BPNN | Neuron numbers in the middle layer |
|---|---|
| 44% | 20 |
| 79.6% | 30 |
| 85% | 40 |
| 98.4% | 50 |
| 98% | 60 |

As it could be seen from the above table 1, the best result is obtained from using 50 neurons in the middle layer and from that on, the increase in neuron numbers in this layer does not enhance the system precision. Only the number of calculations increases.

To make this system parallel by CUDA, a GeForce GT 430 graphic card is used which has 96 cores. As mentioned before, this paralleling is done in all the four steps of BPNN and finally its obtained results, running order ratio which is run on its processor 'Core (TM) i7 CPU 3.2 GHz' are shown in the table 2:

**Table 2.**

| processor | Run time (second) | Speed enhancement |
|---|---|---|
| CPU | 3521 | |
| GPU | 96 | **36** |

## 5. Conclusion and future works

As seen in the previous section, we could enhance the speed of iris recognition to 45 times higher by using parallel algorithm and CUDA to implement it. The interesting point is that this speeds enhancement is gained through the ordinary graphic card. In case that a stronger and multi-core graphic card could be used, this enhancement would be much higher.

Another interesting point is that in the real situation applications we commonly encounter with voluminous databases, thus the time matching pattern is much higher in these systems. But, CUDA could not be used to solve this problem since the memory in the graphic card is limited and the whole databases could not be loaded on it. To overcome this problem, cluster could be used in a way that we can distribute the databases on cluster nodes. While an input data is received, a copy of it is sent to all the nodes and every node performs the recognition process and finally the results will be sent to the PC source.